\documentclass[11pt]{article}
\usepackage[letterpaper,margin=0.85in]{geometry}
\usepackage{amsmath,amssymb}
\usepackage{booktabs}
\usepackage{float}
\usepackage{caption}
\usepackage{pgfplots}
\pgfplotsset{compat=1.17, every axis/.append style={cycle list name=black white}}
\usepackage[hidelinks]{hyperref}
\usepackage{authblk}
\usepackage[T1]{fontenc}
\usepackage{lmodern}
\usepackage[expansion=false]{microtype}
\title{\vspace{-2.2em}
\Large\textbf{Stranded Compute in AI Inference Workloads:}\\[0.15em]
\textbf{Recovering Value Through Demand Aggregation} \\

\vspace{-0.3em}}
\author{Sam Kingsley$^{1}$}

\affil{$^{1}$Haas School of Business, University of California, Berkeley\\
\small\texttt{sam\_kingsley@berkeley.edu}}
\date{\small August 2026}

\begin{document}
\maketitle
\vspace{-2.4em}

\begin{abstract}
\noindent\small
The AI inference infrastructure is commonly analyzed as an energy business. Under full total-cost accounting it is
not: in the reference deployment modeled here, capital recovery is $87.3\%$ of fully loaded cost per token
and electricity $3.7\%$. Because capital accrues per unit time irrespective of use, accelerator value is
destroyed by non-utilization rather than by consumption. Yet, SLO-qualified goodput reaches only $59.6\%$ of
nameplate throughput, and the decomposition of the loss chain shows that $25.2$ of the $40.4$ lost percentage
points arise from scheduling idleness and stranded burst reserve. Both are properties of a tenant's demand
portfolio rather than of its silicon, and neither is diversifiable by a single tenant acting alone. Modelling
reserve and idle capacity as diversifiable risk, pooling demand across eight imperfectly correlated tenants
($\rho=0.3$) recovers $7.3$ percentage points of goodput, approximately \$0.75M per annum per megawatt of IT
capacity, or between $2.8\times$ and $4.9\times$ the entire annual interconnection spend of the deployment that
produced it. The implication for neutral colocation and interconnection providers is that the switching and interconnection layer aggregating tenant demand functions as an instrument of capital productivity while being priced as transport.
\end{abstract}

\vspace{-0.6em}
\section{Introduction}

The prevailing model for AI data-centre economics is an energy model. Capacity is quoted in megawatts,
efficiency in PUE, and the binding constraint described as time-to-power. That framing is correct about the
\emph{physical} constraint and misleading about the \emph{economic} one. Once accelerator capital is amortised
honestly---over a four-year ownership horizon, inclusive of financing carry \cite{semianalysis2026}---inference
cost is dominated by capital recovery to a degree that inverts the standard intuition about where margin is
won and lost.

This matters because capital recovery is invariant to utilisation. A megawatt-hour not consumed is not paid
for; a GPU-hour not sold is gone. When the invariant component is $87\%$ of cost, the economics collapse onto
one variable: the fraction of installed throughput converted into billable, SLO-compliant output.

The question here is deliberately narrow. Given capital dominance, and given that observed goodput falls well
short of nameplate, \emph{which components of the shortfall are addressable, by what mechanism, and what is
recovery worth relative to the cost of the mechanism?} The answer developed below is that the largest
addressable losses are statistical rather than technical, that their remedy requires demand from multiple
uncorrelated sources, and that the interconnection layer is the practical instrument of that aggregation.\\

The analysis is developed for a representative carrier-neutral colocation deployment; the mechanism is general to any multi-tenant facility with interconnection capability.
\newpage
\section{Physical and Systems Constraints}

Three physical facts establish the economic structure.

\textbf{Idle accelerators consume power without producing revenue.} Measured GPU energy decomposes into a
fixed request-level term and a marginal per-token term, $E(N)=E_{\mathrm{fixed}}+N\,E_{\mathrm{step}}$
\cite{vellaisamy2026}. Attribution to a resident floor component gives a median floor share of $51.6\%$ of request energy, rising to
$79.1\%$ in low-load short-decode regimes \cite{vellaisamy2026}. An under-utilised accelerator burns most of
its energy producing nothing, compounding capital loss rather than offsetting it.

\textbf{Throughput cannot be pushed arbitrarily to fill idle capacity.} Serving sits on a Pareto frontier
between cost per token and serial generation speed \cite{erdil2025}. An optimal batch size $b^*$ exists at which parameter-read time equals arithmetic time; beyond it, larger
batches raise latency without reducing GPU-seconds per token, and past an optimal instance size all-reduce
latency overwhelms added parallelism \cite{erdil2025}. Idle capacity cannot simply be absorbed by running
existing requests harder.

\textbf{Capacity must be held in reserve to meet service levels.} Interactive inference carries tail-latency
commitments. Meeting a p99 target under bursty arrivals requires memory and compute headroom that is, by
construction, unsold in normal operation. This is not engineering waste---it is the price of the service
level---but it is capital generating no revenue.

\section{Economic Model}

\subsection{The goodput identity}
Billable output is nameplate throughput net of a multiplicative chain of losses:
\begin{equation}
G \;=\; N \cdot u \cdot (1-r) \cdot (1-f) \cdot (1-m) \cdot s \cdot q
\label{eq:goodput}
\end{equation}
with $G$ tokens per hour (rate), $N$ nameplate throughput, $u$ scheduler utilisation, $r$ burst reserve, $f$ fragmentation and low-batch
loss, $m$ maintenance downtime, $s$ SLO compliance, $q$ quality acceptance. The factors compound: six
individually unremarkable losses reduce billable output to $59.6\%$ of nameplate.

\subsection{Why utilisation is the only variable that matters}
Let fully loaded annual cost be $C=C_{\mathrm{cap}}+C_{\mathrm{op}}$ and revenue $R=G H p$ for $H$ operating
hours and price $p$ per token. Since $C_{\mathrm{cap}}$ is invariant in $G$ and electricity is the only
materially variable cost,
\begin{equation}
\frac{\partial \Pi}{\partial G}\;=\;Hp-\frac{\partial C_{\mathrm{op}}}{\partial G}
\;\approx\;Hp\!\left(1-\frac{C_{\mathrm{elec}}}{R}\right)
\label{eq:marginal}
\end{equation}
With electricity at $3.7\%$ of fully loaded cost, roughly $88$ cents of each incremental revenue dollar falls
to gross profit. Symmetrically, each point of goodput not recovered is a near-total loss of its assigned capital. $G \cdot H$ denotes the annual quantity of billable tokens produced, measured in tokens/year.

\subsection{Reserve and idle as diversifiable risk}
Burst reserve absorbs variance in arrival rate. For $n$ tenants with pairwise demand correlation $\rho$, the
standard deviation of aggregate demand per unit capacity scales as
\begin{equation}
\sigma_n=\sigma_1\sqrt{\tfrac{1+(n-1)\rho}{n}},\qquad
r_n=r_1\sqrt{\tfrac{1+(n-1)\rho}{n}},\qquad
1-u_n=(1-u_1)\sqrt{\tfrac{1+(n-1)\rho}{n}}
\label{eq:pool}
\end{equation}
At $\rho=0$ this reduces to the classical $1/\sqrt{n}$ result; at $\rho=1$ pooling yields nothing. The limit as
$n\!\to\!\infty$ is $\sigma_1\sqrt{\rho}$: diversification has a floor set by correlation, not by scale.

This is the paper's central structural claim. \emph{Two of the six loss terms in Equation~\ref{eq:goodput} are
not properties of the silicon or the serving stack. They are properties of the demand portfolio.} A tenant
cannot diversify its own arrival process; uncorrelated demand is an external input.

\section{Unit Economics}

\begin{table}[H]
\centering\small
\caption{Key model inputs. Provenance: \textsc{s} sourced, \textsc{a} assumed.}
\label{tab:assumptions}
\begin{tabular}{@{}llc@{\hspace{2.2em}}llc@{}}
\toprule
Input & Value & Prov. & Input & Value & Prov. \\
\midrule
IT power budget & 1{,}000 kW & \textsc{a} & Electricity price & \$0.07/kWh & \textsc{s} \\
Accelerators (10 racks $\times$ 72) & 720 & \textsc{a} & PUE (liquid-cooled) & 1.15 & \textsc{s} \\
Accelerator rack capital & \$3.90M/rack & \textsc{a} & Token price & \$0.50/M tok & \textsc{a} \\
Facility capital & \$11{,}000/kW & \textsc{a} & Scheduler utilisation $u$ & 85\% & \textsc{a} \\
Network + storage capital & \$1{,}800/kW & \textsc{a} & Burst reserve $r$ & 12\% & \textsc{a} \\
Economic life / residual & 4 yr / 15\% & \textsc{a} & Fragmentation $f$ / maint.\ $m$ & 6\% / 3\% & \textsc{a} \\
Cost of capital & 11.0\% & \textsc{a} & SLO $s$ / quality $q$ & 92\% / 95\% & \textsc{a} \\
\bottomrule
\end{tabular}
\end{table}

\vspace{-0.6em}
\begin{table}[H]
\centering\small
\caption{Unit economics, calculated from Table~\ref{tab:assumptions} via Eqs.~\ref{eq:goodput}--\ref{eq:marginal}.}
\label{tab:unit}
\begin{tabular}{@{}lr@{\hspace{2.6em}}lr@{}}
\toprule
Cost structure & Value & Throughput and margin & Value \\
\midrule
Deployed capital & \$51.8M & Nameplate throughput & 2.35 Gtok/hr \\
Annual capital recovery & \$16.71M & \textbf{SLO-qualified goodput} & \textbf{59.6\% of nameplate} \\
Annual operating cost & \$2.43M & Overstatement if goodput ignored & 42.6\% \\
Fully loaded annual cost & \$19.14M & Fully loaded cost / M tokens & \$1.56 \\
\textbf{Capital share of cost} & \textbf{87.3\%} & Revenue per facility MW & \$5.34M \\
\textbf{Electricity share of cost} & \textbf{3.7\%} & \textbf{Value of 1 pp of goodput} & \textbf{\$102.9k/yr/MW} \\
Labour, software, water share & 9.0\% & \quad net of marginal electricity & \$91.1k/yr/MW \\
\bottomrule
\end{tabular}
\end{table}

Two items are worth highlighting. Accelerator rack capital is an estimate, not a
sourced quotation; it drives $87\%$ of the cost stack and most deserves replacement with a real vendor price.
And at the default token price the deployment does not clear its cost of capital---fully loaded cost is
\$1.56/M against an assumed \$0.50/M. That is itself a finding. Results below use the default price and are
therefore conservative; at \$1.56/M one recovered point is worth roughly \$321k per annum per MW.

\vspace{-0.3em}
\section{Where the Losses Actually Are}

Of $40.4$ percentage points of total loss, idle capacity accounts for $15.0$ and burst reserve for $10.2$---
together $25.2$ points, or $62\%$ of all loss. The remainder (fragmentation $4.5$, maintenance $2.1$, SLO
misses $5.5$, quality rejection $3.1$) is addressable by serving-stack work \cite{kwon2023}, operational
reliability, and model routing. Those are real problems with identifiable engineering owners. The two largest
are not.

\begin{figure}[H]
\centering
\begin{tikzpicture}
\begin{axis}[
  width=14.2cm, height=4.3cm, ybar, bar width=26pt,
  ymin=0, ymax=115, ylabel={\scriptsize \% of nameplate},
  symbolic x coords={Nameplate,Idle,Reserve,Frag.,Maint.,SLO miss,Quality,Goodput},
  xtick=data, x tick label style={font=\scriptsize}, ytick={0,25,50,75,100},
  tick label style={font=\scriptsize}, ylabel style={font=\scriptsize},
  nodes near coords, nodes near coords style={font=\tiny},
  enlarge x limits=0.07, every axis plot/.append style={fill=black!14,draw=black},
]
\addplot[fill=black!15,draw=black] coordinates {(Nameplate,100) (Idle,15.0) (Reserve,10.2) (Frag.,4.5) (Maint.,2.1) (SLO miss,5.5) (Quality,3.1) (Goodput,59.6)};
\end{axis}
\end{tikzpicture}
\caption{Decomposition of the goodput identity. Idle and burst reserve account for $25.2$ of the $40.4$-point
shortfall. Both are properties of the demand portfolio rather than of silicon or serving stack, and neither is
diversifiable by a single tenant acting alone.}
\label{fig:waterfall}
\end{figure}
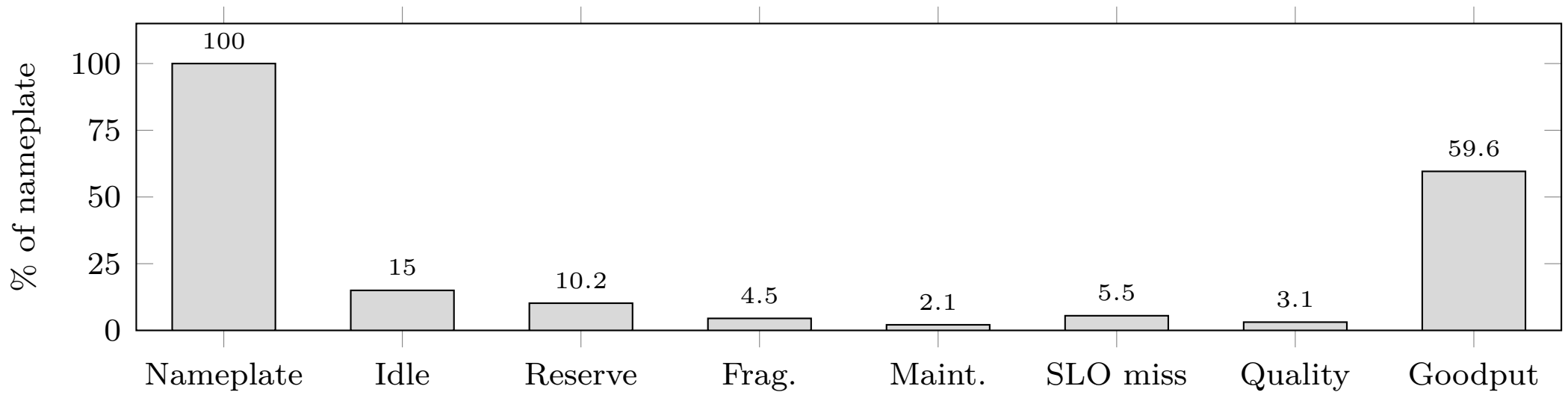

\vspace{-0.6em}
\section{The Colocation Mechanism}

If the two largest losses diversify with the number of uncorrelated demand sources, the economic question becomes: what physical mechanism delivers uncorrelated demand to installed capacity? A carrier-neutral colocation facility is such a mechanism. By co-locating tenants whose demand is driven by different end markets, time zones, and product cycles, it assembles the multi-source demand portfolio that no single operator can construct alone. The interconnection layer within that facility, cross-connects, virtual circuits, exchange points, is not, under this framing, a transport service. It is the switching fabric that pools demand streams, and its economic output is recovered capital productivity in the inference deployments it connects.

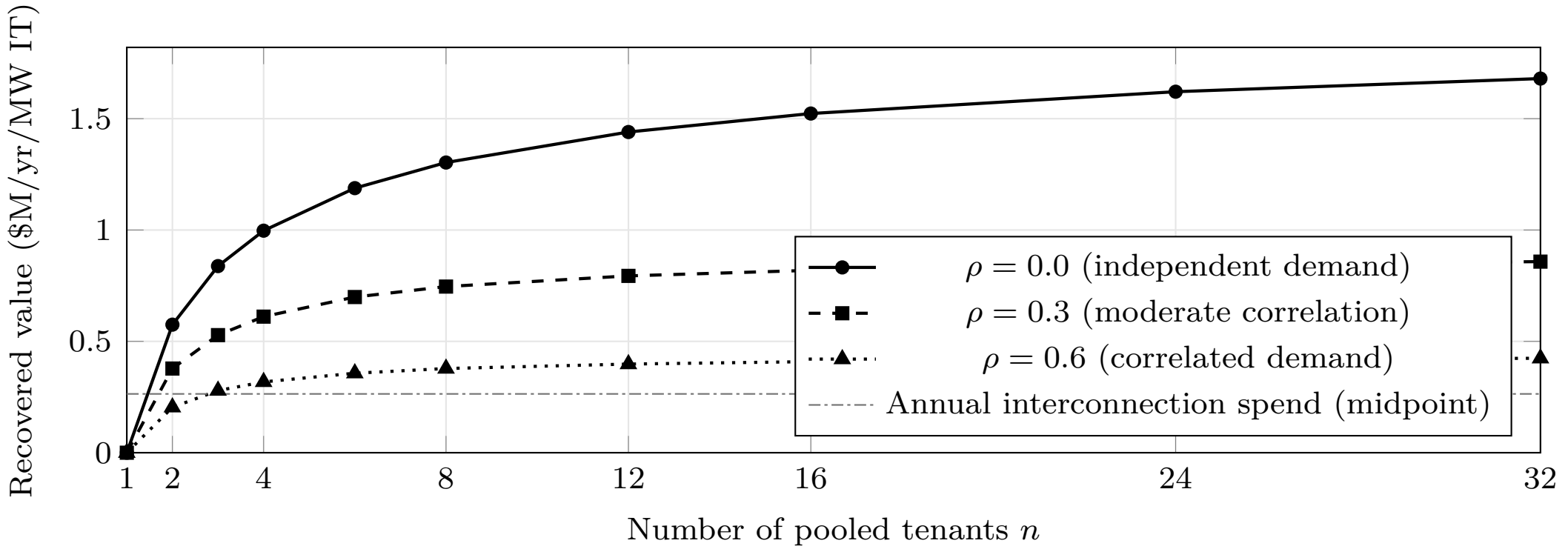
\begin{figure}[H]
\centering
\begin{tikzpicture}
\begin{axis}[
  width=14.2cm, height=5.2cm,
  xlabel={\scriptsize Number of pooled tenants $n$},
  ylabel={\scriptsize Recovered value (\$M/yr/MW IT)},
  xmin=1, xmax=32, ymin=0, ymax=1.82, xtick={1,2,4,8,12,16,24,32},
  tick label style={font=\scriptsize}, xlabel style={font=\scriptsize}, ylabel style={font=\scriptsize},
  legend style={font=\scriptsize, at={(0.98,0.04)}, anchor=south east, draw=black},
  grid=major, grid style={black!10}, every axis plot/.append style={thick, black},
]
\addplot[black,mark=*,mark size=1.4pt,mark options={fill=black}] coordinates {(1,0)(2,.575)(3,.838)(4,.997)(6,1.188)(8,1.303)(12,1.440)(16,1.523)(24,1.621)(32,1.680)};
\addlegendentry{$\rho=0.0$ (independent demand)}
\addplot[black,mark=square*,mark size=1.4pt,dashed,mark options={fill=black,solid}] coordinates {(1,0)(2,.378)(3,.528)(4,.611)(6,.699)(8,.746)(12,.794)(16,.819)(24,.845)(32,.858)};
\addlegendentry{$\rho=0.3$ (moderate correlation)}
\addplot[black,mark=triangle*,mark size=1.8pt,dotted,mark options={fill=black,solid}] coordinates {(1,0)(2,.204)(3,.279)(4,.318)(6,.357)(8,.378)(12,.398)(16,.409)(24,.419)(32,.424)};
\addlegendentry{$\rho=0.6$ (correlated demand)}
\addplot[black!55,mark=none,densely dashdotted,thin] coordinates {(1,0.264)(32,0.264)};
\addlegendentry{Annual interconnection spend (midpoint)}
\end{axis}
\end{tikzpicture}
\caption{Value of goodput recovered through demand pooling, by tenant count and demand correlation, at the
default token price. Diversification saturates quickly---most benefit is captured by roughly eight
tenants---and its ceiling is set by correlation rather than tenant count.}
\label{fig:pooling}
\end{figure}

Figure~\ref{fig:pooling} quantifies the recovery. At eight pooled tenants with moderate correlation
($\rho=0.3$), goodput rises from $59.6\%$ to $66.9\%$ of nameplate, worth \$0.75M per annum per MW of IT
capacity; under weak correlation, \$1.30M. The reference line is the midpoint of estimated annual
interconnection spend for a deployment of this size---on the order of \$132k to \$528k, assuming 20--80
cross-connects and 2--8 high-capacity virtual ports at prevailing list rates (illustrative).

\vspace{-0.4em}
The ratio is the result. At moderate correlation and eight tenants, recovered capital productivity is roughly
$2.8\times$ the annual interconnection bill; under weak correlation, $4.9\times$. The interconnection layer
captures a fraction of the value it enables, because it is priced per port, per circuit and per cross-connect
rather than as a function of the capital productivity it restores.

Two secondary implications follow without additional assumptions. Diversification saturates, so the relevant
unit of value is the \emph{diversity} of a metro's demand portfolio rather than its raw tenant count. And
because the ceiling is $\sigma_1\sqrt{\rho}$, a platform whose tenants span genuinely different end markets is
worth materially more than an equally large one concentrated in a single vertical---an argument for
composition over scale.

\vspace{-0.3em}
\section{Sensitivity}

\begin{table}[H]
\centering\small
\caption{Recovered value at $n=8$ pooled tenants, \$M per annum per MW IT, and recovery multiples against the
interconnection-spend range.}
\label{tab:sens}
\begin{tabular}{@{}lccc@{}}
\toprule
& $\rho=0.0$ & $\rho=0.3$ & $\rho=0.6$ \\
\midrule
Token price \$0.25/M & 0.65 & 0.37 & 0.19 \\
Token price \$0.50/M (default) & 1.30 & 0.75 & 0.38 \\
Token price \$1.56/M (break-even) & 4.06 & 2.33 & 1.18 \\
\midrule
Multiple vs.\ low interconnection spend (\$0.13M), at \$0.50/M & 9.9$\times$ & 5.6$\times$ & 2.9$\times$ \\
Multiple vs.\ high interconnection spend (\$0.53M), at \$0.50/M & 2.5$\times$ & 1.4$\times$ & 0.7$\times$ \\
\bottomrule
\end{tabular}
\end{table}

\vspace{-0.4em}
Token price enters linearly and is the dominant sensitivity. The conclusion reverses only in the corner where
demand is both highly correlated ($\rho\geq0.6$) and interconnection spend is at the high end---that is, where
a platform sells a great deal of connectivity to tenants whose demand moves together. That corner is
diagnostic rather than fatal: it identifies precisely the customer portfolio for which the mechanism does not apply.

\vspace{-0.3em}
\section{Conclusion}

Under full total-cost accounting, inference is a capital business in which $87\%$ of cost is invariant to use.
That makes utilisation the dominant economic variable and every unrecovered point of goodput a near-total
capital loss. The majority of the shortfall---$25.2$ of $40.4$ percentage points---consists of scheduling
idleness and stranded burst reserve, both properties of a tenant's demand portfolio rather than of its
hardware or software. Such problems are diversifiable, but not by the tenant that owns them: they require
access to uncorrelated demand, which an interconnection platform structurally provides. Pooling across eight
imperfectly correlated tenants recovers value on the order of \$0.75M per annum per megawatt---several
multiples of the interconnection spend of the deployment generating it. Whether that gap represents an
opportunity for repricing, for product design, or for nothing at all is a commercial judgement. The economic
observation is that it exists, that it follows from the physics of capital-intensive infrastructure rather
than from any claim about AI adoption, and that it is currently invisible in how interconnection is priced across the carrier-neutral colocation market.


\vspace{-0.4em}
\begin{thebibliography}{9}\small
\bibitem{erdil2025} E.~Erdil. Inference economics of language models. \emph{arXiv:2506.04645}, 2025.
\bibitem{vellaisamy2026} P.~Vellaisamy, V.~Lam, S.~Blanton, J.~P.~Shen. Characterization of request and token
energy costs for LLM inference workloads on GPU platforms. \emph{IISWC 2026}; \emph{arXiv:2608.28044}.
\bibitem{peng2026} S.-W.~Peng, Y.-H.~Lin, Y.-P.~Lee. Inference economics of enterprise coding agents: cloud vs.\
on-premise LLMs. \emph{arXiv:2607.13080}, 2026. Source for utilisation-dependence of per-token cost.
\bibitem{kwon2023} W.~Kwon et al. Efficient memory management for LLM serving with PagedAttention.
\emph{SOSP 2023}; \emph{arXiv:2309.06180}.
\bibitem{semianalysis2026} SemiAnalysis. InferenceMAX: open-source inference benchmarking, 2026. Four-year
ownership horizon adopted here for TCO.
\bibitem{uptime2023} Uptime Institute. Global Data Center Survey 2023. Source for PUE reference values.
\end{thebibliography}
\end{document}